\documentclass[sn-mathphys-num]{sn-jnl}

\usepackage{graphicx}%
\usepackage{multirow}%
\usepackage{amsmath,amssymb,amsfonts}%
\usepackage{amsthm}%
\usepackage{mathrsfs}%
\usepackage[title]{appendix}%
\usepackage{xcolor}%
\usepackage{textcomp}%
\usepackage{manyfoot}%
\usepackage{booktabs}%
\usepackage{algorithm}%
\usepackage{algorithmicx}%
\usepackage{algpseudocode}%
\usepackage{listings}%
\usepackage{subcaption}
\usepackage{comment}

\theoremstyle{thmstyleone}%
\newtheorem{theorem}{Theorem}
\theoremstyle{thmstyletwo}%

\theoremstyle{thmstylethree}%

\begin{document}


\title[Article Title]{Robust Node Affinities via Jaccard-Biased Random Walks and Rank Aggregation}


\author*[1]{\fnm{Bastian} \sur{Pfeifer}}\email{bastian.pfeifer@medunigraz.at}


\author[1]{\fnm{Michael G.} \sur{Schimek}}\email{michael.schimek@medunigraz.at}

\affil*[1]{
Institute for Medical Informatics, Statistics and Documentation, Medical University Graz, Austria}



\abstract{Estimating node similarity is a fundamental task in network analysis and graph-based machine learning, with applications in clustering, community detection, classification, and recommendation. We propose TopKGraphs, a method based on start-node–anchored random walks that bias transitions toward nodes with structurally similar neighborhoods, measured via Jaccard similarity. Rather than computing stationary distributions, walks are treated as stochastic neighborhood samplers, producing partial node rankings that are aggregated using robust rank aggregation to construct interpretable node-to-node affinity matrices.

TopKGraphs provides a non-parametric, interpretable, and general-purpose representation of node similarity that can be applied in both network analysis and machine learning workflows. We evaluate the method on synthetic graphs (stochastic block models, Lancichinetti-Fortunato-Radicchi benchmark graphs), k-nearest-neighbor graphs from tabular datasets, and a curated high-confidence protein–protein interaction network. Across all scenarios, TopKGraphs achieves competitive or superior performance compared to standard similarity measures (Jaccard, Dice), a diffusion-based method (personalized PageRank), and an embedding-based approach (Node2Vec), demonstrating robustness in sparse, noisy, or heterogeneous networks.

These results suggest that TopKGraphs is a versatile and interpretable tool for bridging simple local similarity measures with more complex embedding-based approaches, facilitating both data mining and network analysis applications.}

\keywords{networks, node affinity, random walk, rank aggregation, clustering}



\maketitle

\section{Introduction}
\label{intro}

Networks provide a flexible and powerful framework for representing complex systems in many domains, including biology, medicine, social sciences, and engineering \cite{newman2010networks, barabasi2016network}. In a network, nodes represent entities (e.g., genes, proteins, individuals, or components), while edges encode interactions, relationships, or associations between them.

Random walks are a fundamental tool for exploring these networks, capturing both local and global connectivity patterns and enabling quantitative measures of node similarity, influence, or importance \cite{lovasz1993random, masuda2017random, cao2022graph}. By simulating stochastic traversals over the graph, random walks allow information to propagate beyond immediate neighbors. Accurately quantifying structural similarity between nodes is a central task in graph-based machine learning, with applications in node classification, clustering, community detection, recommendation, and link prediction.

Biomedical networks, including molecular interaction, patient similarity, and drug–disease networks, provide a concrete example where node similarity is crucial. In these networks, shared interactions or neighborhoods often reflect functional or clinical relatedness. Simple set-based metrics such as Jaccard similarity are widely used due to their interpretability and robustness to sparse data, serving as benchmarks or building blocks for more complex predictive models in applications ranging from disease module discovery to drug repurposing \cite{polanco2025drug, hu2026enhancing, zitnik2024current, pfeifer2022gnn, pfeifer2023parea}.

In this work, we introduce TopKGraphs, a method for computing interpretable node-to-node affinity matrices using start-node–anchored random walks combined with robust rank aggregation. Each walk is biased toward nodes with similar local neighborhoods, measured via Jaccard similarity, and first-visit orderings are aggregated across walks using a penalized Borda mean \cite{borda1781memoire, schimek2015topklists}. The resulting matrices capture both local overlap and multi-hop structural context while remaining parameter-light, interpretable, and suitable for integration into downstream graph learning tasks such as node classification and clustering.

The remainder of this manuscript is organized as follows. In Section~\ref{sec:related}, we review relevant work on random-walk embeddings, ranking, and hybrid graph methods. We highlight the distinctions and motivations behind TopKGraphs. Section~\ref{sec:method} presents the TopKGraphs methodology. We detail the start-node–anchored random walk, the construction of node-specific affinity matrices via first-visit order, and the Borda-based aggregation procedure. Section~\ref{sec:evaluation} describes our comprehensive evaluation strategy. This includes synthetic benchmarks based on stochastic block models (SBM) and Lancichinetti-Fortunato-Radicchi (LFR) \cite{lancichinetti2008benchmark} graphs, as well as real-world networks such as k-nearest-neighbor graphs retrieved from tabular data, the CORA citation network, and a curated protein–protein interaction network. Section~\ref{sec:results} presents results on community detection and node classification. We highlight the robustness to random walk parameters, computational efficiency, and comparative performance against established baseline methods. Finally, Section~\ref{sec:conclusion} discusses limitations and directions for future work.

\section{Related Work}
\label{sec:related}

Our algorithm builds on the rich literature of random‑walk–based embedding and ranking methods, but introduces a novel approach to biasing walks and aggregating information, with a focus on parameter efficiency and interpretability.

\textit{Random‑walk embeddings.} Methods such as DeepWalk \cite{perozzi2014deepwalk} and Node2Vec \cite{grover2016node2vec} generate sequences of nodes via random walks and feed them into skip‑gram–style models to learn continuous node embeddings for downstream tasks. These methods rely on several hyperparameters, including walk length, number of walks per node, skip‑gram window size, embedding dimension, and, in the case of Node2Vec, two additional parameters ($p$ and $q$). The parameter \(p\) controls the likelihood of immediately revisiting a node (encouraging or discouraging backtracking), while \(q\) biases the walk toward breadth-first (higher \(q\)) or depth-first (lower \(q\)) exploration, allowing a trade-off between local and global network structure in the embeddings.

Extensions include role‑aware random walk methods and structural embedding techniques like Struc2Vec that emphasize structural identity rather than proximity alone \cite{ribeiro2017struc2vec}. While powerful, the need to tune many parameters can be cumbersome, especially in unsupervised settings with limited supervision.

\textit{Random‑walk–based ranking.} Personalized PageRank (PPR) \cite{page1999pagerank,gleich2015pagerank} ranks nodes around a seed set using a random walk with restarts, aggregating landing probabilities over all walk lengths. Variants such as rooted PageRank and asymmetric proximity‑preserving embeddings exploit personalized restart dynamics and random walk diffusion to capture directed or high‑order similarities \cite{zhou2017app}. Extensions of PageRank have also been adapted for context‑aware recommendation and proximity modeling in heterogeneous graphs \cite{musto2021contextpagerank}. These methods also require parameter choices, such as the restart probability, which can influence performance and may need tuning for different datasets. In contrast, TopKGraphs differs fundamentally from diffusion-based similarities such as PageRank or heat kernels, which aggregate stationary or time-averaged visitation probabilities. Instead, it treats random walks as stochastic samplers of ranked neighborhood structure, discarding frequency information in favor of first-visit order and rank aggregation. This design emphasizes relative proximity and stability over precise diffusion mass.

\textit{Random walks in GNNs and hybrid models.} Recent work has explored integrating random walk and PageRank principles into graph neural network architectures to improve structural awareness and mitigate common issues such as oversmoothing. For example, random‑walk aggregation mechanisms augment traditional message passing by capturing path‑based neighborhoods \cite{jin2022raw}, while hierarchical PageRank can be used to inform multi‑scale neighborhood aggregation in GNNs \cite{yang2023shpgnn}. Personalized PageRank has also been adapted to define infinitely deep neural architectures that maintain locality and avoid oversmoothing, yielding improved performance on a range of tasks \cite{roth2022pprgnn}. Additional recent models integrate personalized or higher‑order PageRank directly into GNN convolution operations to better capture long‑range and heterophilic interactions \cite{wang2025hpgnn}.

\textit{Our approach.} Unlike the aforementioned methods, our algorithm defines a first‑order, start‑node–specific biased random walk in which transition probabilities are determined by one‑hop Jaccard similarity to the source node, rather than by node degree, generic restart probabilities, or second‑order memory. Importantly, it only requires two interpretable parameters: the number of walks and the walk length. This simplicity is especially beneficial in unsupervised settings, where labeled data is scarce and extensive hyperparameter tuning is impractical. Instead of training embeddings on co‑occurrence windows or relying on stationary distributions, we interpret entire walk trajectories via first‑visit order and aggregate multiple walks into a node‑centric interpretable affinity matrix. 

\section{Method}
\label{sec:method}

Let $G = (V,E)$ be an undirected, unweighted graph with
vertex set $V = \{1,\dots,n\}$ and edge set
$E \subseteq \{(u,v) : u,v \in V,\ u \neq v\}$.
Throughout this section, the terms \emph{node} and \emph{vertex} are used synonymously.

For each node $v \in V$, let its neighborhood be
\[
N(v) = \{u \in V : (u,v) \in E\}.
\]

Given a start node $s \in V$, our goal is to quantify the structural affinity of all other nodes to $s$.
We first define the Jaccard similarity between a node $v$ and the start node $s$ as
\[
J_s(v) = \frac{|N(v) \cap N(s)|}{|N(v) \cup N(s)|},
\]
with $J_s(v) = 0$ if $N(v) \cup N(s) = \varnothing$.

This similarity is computed once with respect to the start node and remains fixed throughout the walk. To propagate this local similarity through the graph, we define a discrete-time stochastic process
$(X_t)_{t \ge 0}$ with initial state $X_0 = s$.
At each step, given the current state $X_t = u$, the next node $v \in N(u)$ is chosen among the neighbors of $u$ with probability
\[
\mathbb{P}(X_{t+1} = v \mid X_t = u)
=
\frac{J_s(v) + \varepsilon}
{\sum_{v \in N(u)} \bigl(J_s(v) + \varepsilon\bigr)},
\]
where $\varepsilon > 0$ ensures strictly positive transition probabilities.

Consider a maximum walk length $T$, and let $(X_0^{(k)}, \dots, X_T^{(k)})$ denote the $k$-th realization of the stochastic process starting at $X_0^{(k)} = s$, for $k = \{1, \dots, K\}$, where $K$ is the total number of walks.
Let $V_s^{(k)} \subseteq V$ denote the set of nodes visited during this walk, and let
\[
t_s^{(k)}(v) = \min \{ t \in \{0,\dots,T\} : X_t^{(k)} = v \}, \quad v \in V_s^{(k)},
\]
be the first-visit time of each visited node.  

The partial ranking of visited nodes is defined by the order of their first visits, with earlier visits receiving high ranks.  Formally, let $\tau_s^{(k)} : V_s^{(k)} \to \{1,\dots,|V_s^{(k)}|\}$ denote the ranking induced by the first-visit times $t_s^{(k)}$ so that

\[
\tau_s^{(k)}(v) < \tau_s^{(k)}(u) \quad \text{if and only if} \quad t_s^{(k)}(v) < t_s^{(k)}(u), \quad \forall v,u \in V_s^{(k)}.
\]

Here, $|V_s^{(k)}|$ denotes the number of distinct nodes visited during the $k$-th walk starting from $s$. Subsequent revisits are ignored, such that $\tau_s^{(k)}$ depends solely on the order in which nodes are first encountered during the walk.

To extend the partial ranking to all nodes, unvisited nodes are appended in random order at the end. 
Formally, for the $k$-th walk the total ranking is
\[
\tilde{\tau}_s^{(k)}(v) =
\begin{cases}
\tau_s^{(k)}(v) &\text{for}\quad v \in V_s^{(k)},\\[1mm]
|V_s^{(k)}| + r_v^{(k)} &\text{for}\quad v \notin V_s^{(k)},
\end{cases}
\]
where $r_v^{(k)}$ is a random permutation of $\{1,\dots,|V \setminus V_s^{(k)}|\}$ for the unvisited nodes. This preserves the relative order of visited nodes while randomizing the placement of unvisited ones.

Repeating this procedure for $K$ independent walks from the same start node $s$ produces a collection of total rankings
\[
\tilde{\mathcal{T}}_s = \{\tilde{\tau}_s^{(1)}, \dots, \tilde{\tau}_s^{(K)}\}.
\]

A consensus ranking across these walks is obtained via Borda aggregation: for each node $v \in V$, the Borda score is the mean of its total-rank positions over all $K$ walks,
\[
B_s(v) = \frac{1}{K} \sum_{k=1}^{K} \tilde{\tau}_s^{(k)}(v),
\]
where $\tilde{\tau}_s^{(k)}(v)$ denotes the rank of $v$ in the $k$-th walk.  
Smaller Borda scores correspond to nodes that are visited earlier and thus indicate stronger structural affinity to the start node $s$.

The Borda scores induce a total ordering of nodes relative to $s$. Repeating this procedure for all start nodes $s \in V$ yields an asymmetric affinity matrix
\[
A \in \mathbb{R}^{n \times n}, \quad A_{sv} = B_s(v),
\]
which can be row-normalized for interpretability:
\[
A_{sv} \leftarrow \frac{A_{sv}}{\max_{u \in V} A_{su}}.
\]
For applications requiring a symmetric similarity measure, the matrix can be symmetrized as
\[
A \leftarrow \frac{1}{2} (A + A^\top).
\]

For downstream applications, the symmetric affinity matrix can be embedded into a low-dimensional Euclidean space using methods such as classical multidimensional scaling (MDS). These embeddings preserve relative node proximities and can be used for tasks such as k-nearest neighbor classification, clustering, or visualization. In particular, 2D projections of the embedding provide intuitive visualizations of the network structure and node relationships, facilitating exploratory analysis and interpretation.

\subsection{Theoretical Motivation}
\label{sec:theory}

We assume the observed graph $G=(V,E)$ is a perturbed observation of an unobserved \emph{latent similarity graph}
$G^\star=(V,E^\star)$, in which edges represent true structural similarity.
The observed graph is obtained from $G^\star$ via independent edge perturbations:
each latent edge $(u,v)\in E^\star$ is removed with probability $1-p$, and each non-edge $(u,v)\notin E^\star$
is added with probability $q$, where $p\in(0,1]$ and $q\ge 0$.
This model captures both missing and spurious connections arising from noise, sampling effects, or measurement error.

On the latent graph $G^\star$, neighborhood overlap provides a sufficient local statistic for structural affinity.
In particular, the latent Jaccard similarity
\[
J_s^\star(v)
=
\frac{|N^\star(s)\cap N^\star(v)|}{|N^\star(s)\cup N^\star(v)|}
\]
reflects the correct similarity between nodes $s$ and $v$.
However, under edge perturbations, the observed Jaccard similarity
\[
J_s(v)
=
\frac{|N(s)\cap N(v)|}{|N(s)\cup N(v)|}
\]
is generally a biased estimator of $J_s^\star(v)$, as the ratio structure is not preserved under random edge deletions
and additions.

To mitigate this bias, we introduce Jaccard-anchored random walks.
Rather than relying on a single local neighborhood estimate, the walk aggregates multiple noisy similarity observations
along graph paths.
Nodes that are reachable from $s$ through many short paths whose intermediate vertices exhibit high similarity to $s$
are visited earlier in expectation.
Thus, the induced first-visit order acts as a path-based estimator of latent Jaccard proximity.

By averaging first-visit rankings across multiple independent walks using Borda aggregation,
the proposed method yields a robust ordering of nodes by latent structural affinity to the start node,
even in the presence of both missing and spurious edges.

\section{Evaluation Strategy}
\label{sec:evaluation}

To evaluate TopKGraphs comprehensively, we employed a multi-faceted approach spanning both synthetic benchmarks and real-world networks. For synthetic tests, we considered two canonical graph models: stochastic block model (SBM) graphs and Lancichinetti-Fortunato-Radicchi (LFR) \cite{lancichinetti2008benchmark} graphs. SBMs allow precise control over intra- and inter-community connection probabilities, enabling systematic variation of community structure, whereas LFR graphs feature heterogeneous degree distributions and realistic community sizes, more closely resembling the topological characteristics of biomedical networks. For both graph types, multiple instances were generated, and node-affinity matrices derived from TopKGraphs were compared against six alternative approaches: Jaccard similarity, Dice similarity, Laplacian embedding, personalized PageRank, and Node2Vec embeddings. We focus on diffusion-based (personalized PageRank) and embedding-based (Node2Vec) baselines, as they represent the dominant paradigms for random-walk–based similarity estimation, whereas TopKGraphs targets a different objective: rank-based affinity estimation. Community detection was performed via hierarchical clustering using Ward’s linkage \cite{ward1963hierarchical} on each affinity matrix, and clustering performance was quantified using Adjusted Rand Index (ARI), Normalized Mutual Information (NMI), and Adjusted Mutual Information (AMI).

For real-world networks, TopKGraphs was evaluated on the UCI Breast Cancer Wisconsin dataset, the CORA citation network, and a curated human protein-protein interaction (PPI) network. For the breast cancer data, k-nearest-neighbor (kNN) graphs were constructed directly from feature vectors. In CORA, random subgraphs of approximately $100$ nodes were sampled from the citation network, with publication topics serving as ground-truth communities. 

For biological validation, we evaluated all node-affinity methods on a PPI network derived from STRING (\url{https://string-db.org}). Only high-confidence interactions (combined STRING score $\geq 990$) were retained to reduce spurious edges. Protein identifiers were mapped to gene symbols, and only genes with curated disease annotations were kept. Disease labels were obtained from DisGeNET (\url{https://www.disgenet.org}) and used exclusively for evaluation, not during affinity construction, ensuring a strictly unsupervised assessment. After filtering, the largest connected component of the resulting PPI graph was extracted to ensure connectivity for random-walk–based methods. This component consisted of $n = 119$ proteins connected by $m = 314$ undirected interactions, yielding an average node degree of $5.28$ and a low edge density of $0.045$. Despite its moderate size, the graph is sparse and degree-heterogeneous, reflecting typical characteristics of experimentally derived PPI networks. To avoid extreme class imbalance, disease categories with fewer than 20 annotated genes were removed, and overly dominant classes were excluded. The final network comprised nine disease-associated gene groups, including Alzheimer’s disease ($n=11$), breast cancer ($n=19$), hematologic cancer ($n=10$), hypertension ($n=9$), idiopathic pulmonary fibrosis ($n=18$), lung cancer ($n=10$), peripheral nervous system neoplasm ($n=7$), stomach cancer ($n=21$), and urinary bladder cancer ($n=14$). These disease annotations served as ground-truth communities for evaluating clustering performance and as labels for k-nearest neighbor (kNN) node classification.

All experiments were repeated 50 times to account for stochasticity in random walks and subgraph sampling. For TopKGraphs and Node2Vec, walk lengths were systematically varied to examine sensitivity to random-walk parameters. Results were aggregated to report mean values and standard errors, providing a comprehensive view of performance across graph types, methods, and parameter settings. 

\section{Results and Discussion}
\label{sec:results}

\subsubsection*{Synthetic SBM graphs: effect of intra- and inter-community density}
\begin{figure}[h]  
    \centering
    \includegraphics[width=0.7\textwidth]{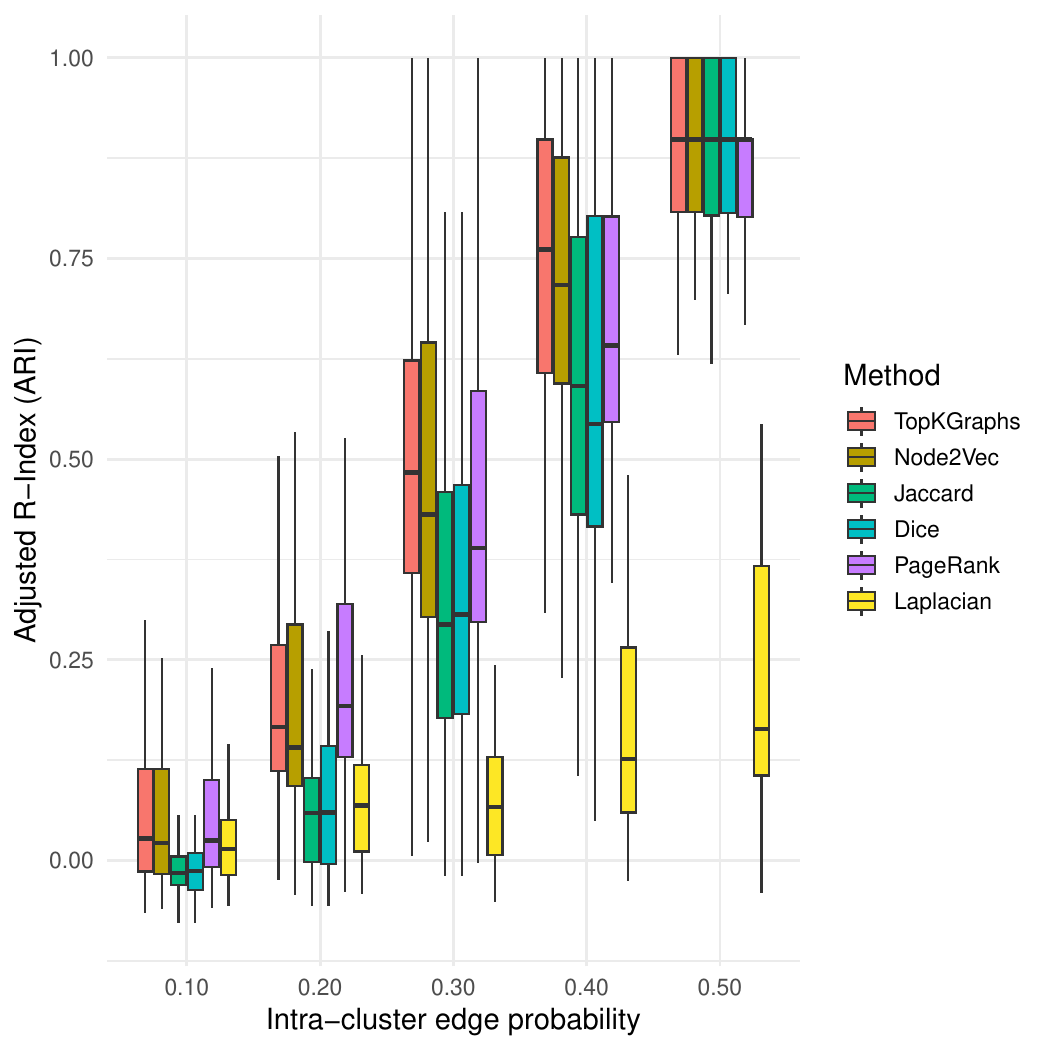}
    \caption{\textbf{Community detection performance on synthetic stochastic block model (SBM) graphs with varying intra-community density.}
Boxplots show the distribution of Adjusted Rand Index (ARI) values across 50 simulations. Hierarchical clustering (Ward’s method) was applied to node affinity matrices derived from six approaches: TopKGraphs, Jaccard similarity, Dice similarity, Laplacian embedding, personalized PageRank, and Node2Vec. Synthetic graphs had three equally sized communities (ten nodes each) with fixed low inter-community connection probability (0.05) and intra-community probabilities ranging from 0.10 to 0.50.}
    \label{fig:intra}
\end{figure}

\begin{figure}[h]  
    \centering
    \includegraphics[width=0.7\textwidth]{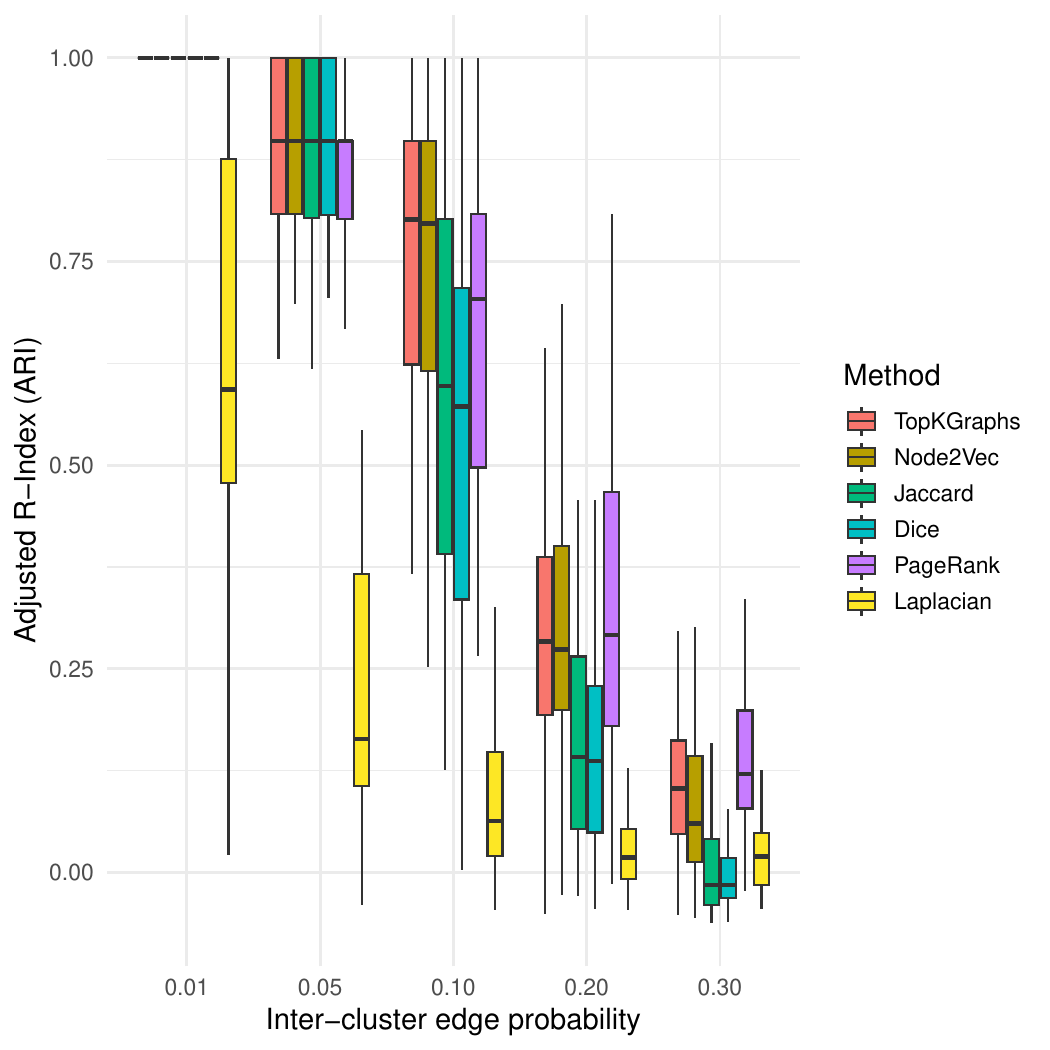}
    \caption{\textbf{Community detection performance on synthetic stochastic block model (SBM) graphs with varying inter-community density.}
Boxplots show the distribution of Adjusted Rand Index (ARI) values across 50 simulations. Hierarchical clustering (Ward’s method) was applied to node affinity matrices derived from six approaches: TopKGraphs, Jaccard similarity, Dice similarity, Laplacian embedding, personalized PageRank, and Node2Vec. Synthetic graphs had three equally sized communities (ten nodes each) with fixed intra-community connection probability of $0.50$ and inter-community probabilities ranging from 0.01 to 0.30.}
    \label{fig:inter}
\end{figure}

On stochastic block model (SBM) graphs TopKGraphs consistently achieved the highest or near-highest Adjusted Rand Index (ARI) across a broad range of intra-community edge probabilities (Fig.~\ref{fig:intra}). As within-community density increased from 0.10 to 0.50 under fixed low inter-community density (0.05), all methods improved, but TopKGraphs and Node2Vec formed the top tier, followed by Jaccard, Dice, and PageRank, with Laplacian embedding trailing. When varying the inter-community edge probability under strong intra-community signal, ARI degraded for all methods as expected (Fig.~\ref{fig:inter}). TopKGraphs maintained a competitive advantage in the moderate-noise regime and remained among the best performers even as inter-cluster density increased, indicating robustness to mixing (see Fig.~\ref{fig:inter}).

\subsubsection*{LFR benchmarks: robustness to mixing and walk-length sensitivity}
\begin{figure}[h]  
    \centering
    \includegraphics[width=0.7\textwidth]{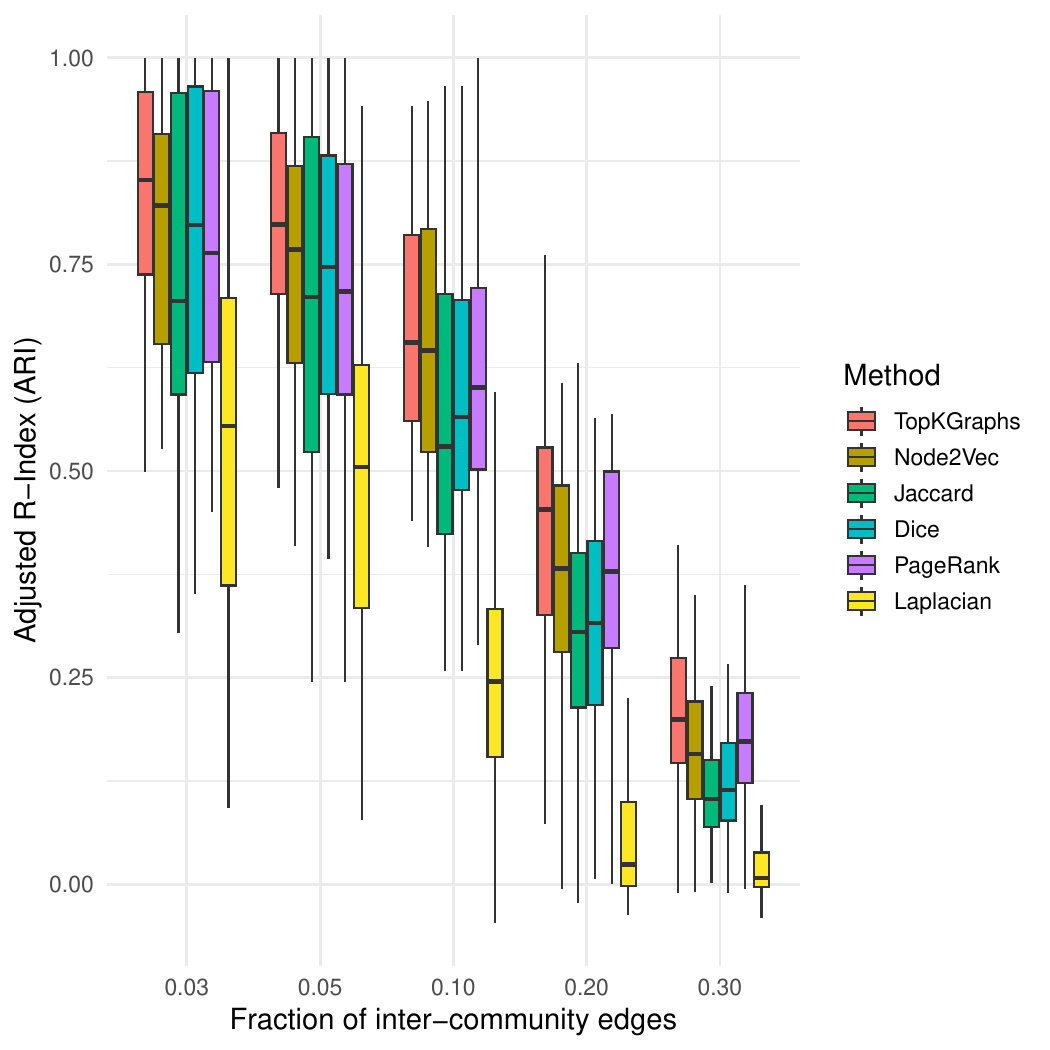}
    \caption{\textbf{Community detection performance on synthetic LFR benchmark graphs.}
Boxplots display the distribution of Adjusted Rand Index (ARI) values over 50 simulation runs. In each iteration, an LFR benchmark graph (100 nodes, average degree 5, maximum degree 10) was generated with varying mixing parameter $\mu = [0.03, 0.05, 0.10, 0.20, 0.30]$, fixed degree exponent $\tau_1 = 2$, and fixed community size exponent $\tau_2 = 1.1$ (community sizes between 5 and 50 nodes). Hierarchical clustering (Ward’s method) was applied to node-affinity matrices derived from six approaches: TopKGraphs, Jaccard similarity, Dice similarity, Laplacian embedding, personalized PageRank, and Node2Vec. Clustering performance was evaluated against the planted LFR community structure.}
    \label{fig:LFR}
\end{figure}

On LFR graphs, performance decreased as the mixing parameter $\mu$ increased, reflecting weaker community structure (Fig.~\ref{fig:LFR}). TopKGraphs and Node2Vec achieved the best results for low to moderate $\mu$, while Jaccard, Dice, and PageRank lagged, and Laplacian embedding dropped most sharply. At higher mixing levels, TopKGraphs consistently outperformed the other methods.

\begin{figure}[htbp]
    \centering
    \begin{subfigure}[b]{0.48\textwidth}
        \centering
        \includegraphics[width=\textwidth]{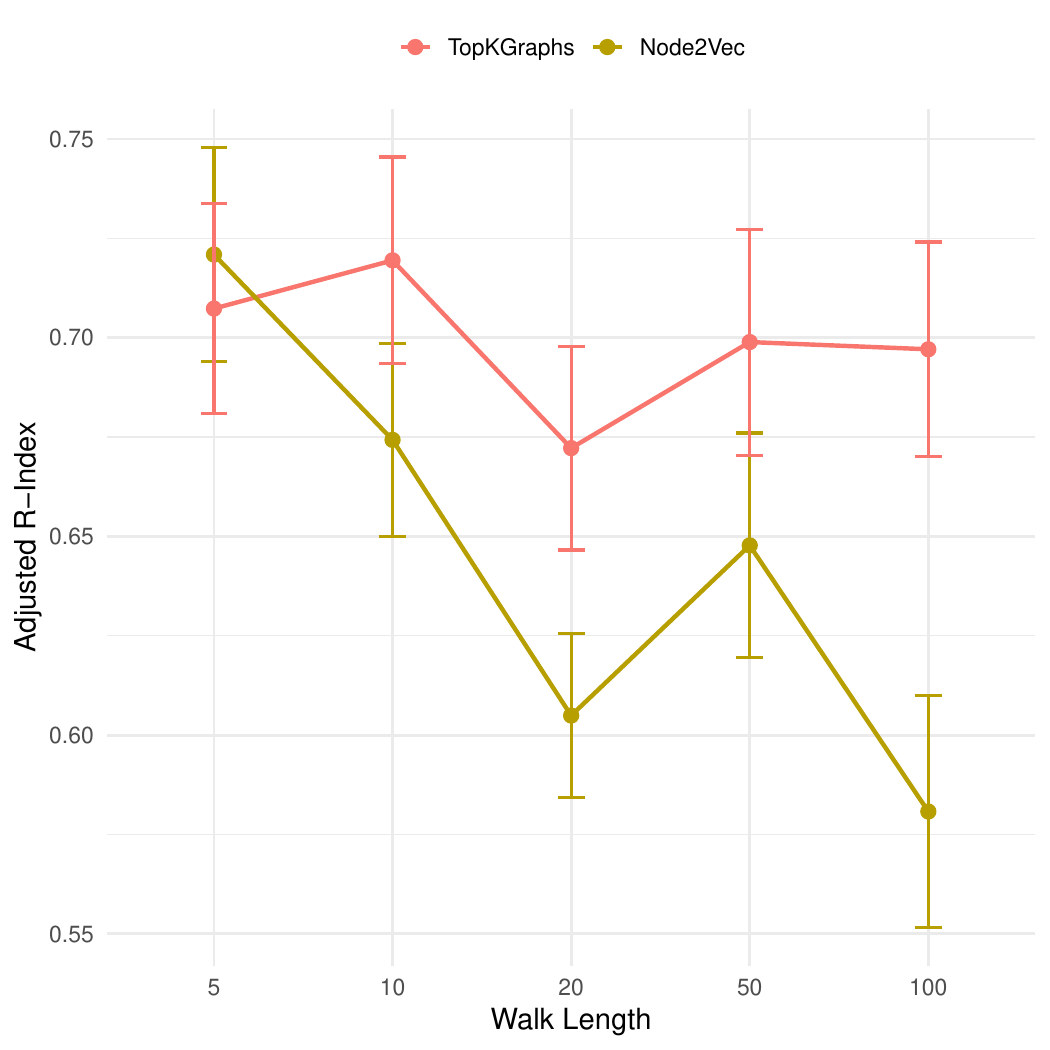}
        \caption{Walk Length}
        \label{fig:LFR_wl}
    \end{subfigure}
    \hfill
    \begin{subfigure}[b]{0.48\textwidth}
        \centering
        \includegraphics[width=\textwidth]{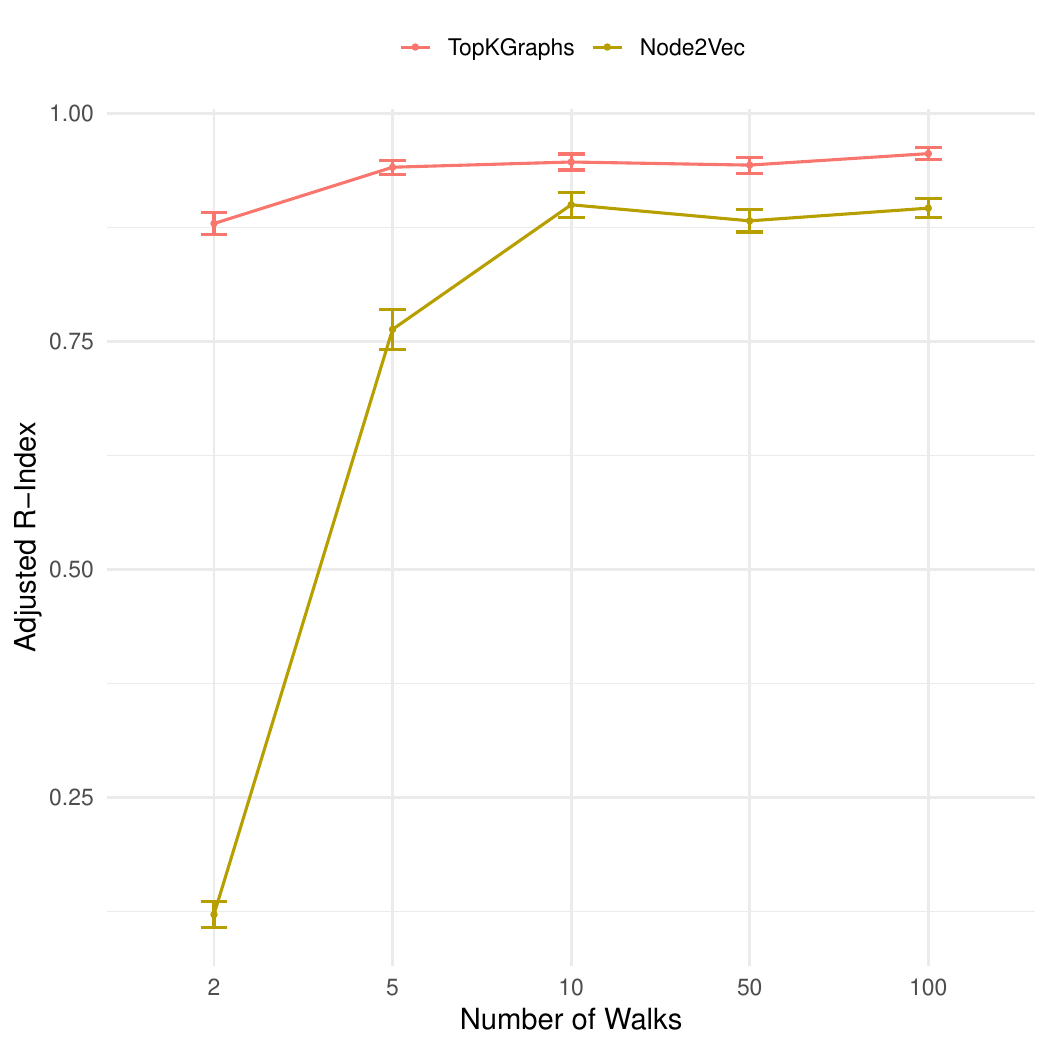}
        \caption{Number of Walks}
        \label{fig:SBM_niter}
    \end{subfigure}
    \caption{\textbf{Effect of random walk length and number of walks on community detection performance.} Mean and standard deviation of Adjusted Rand Index (ARI) values from 50 simulation runs are displayed while subsequently increasing (a) walk length and (b) number of walks. Hierarchical clustering (Ward’s method) was applied to node-affinity representations obtained from TopKGraphs and Node2Vec. (a) LFR benchmark graphs (100 nodes, average degree 5, maximum degree 10, community sizes between 5 and 50 nodes). The mixing parameter was set to $\mu = 0.05$, with degree exponent $\tau_1 = 2$ and community size exponent $\tau_2 = 1.1$. (b) SBM benchmark graphs with three equally sized communities (ten nodes each), fixed intra-community connection probability of $0.50$ and inter-community probabilities of $0.05$.
    }
    \label{fig:fig:LFR_wl_and_SBM_niter}
\end{figure}

Varying the parameter walk length showed that TopKGraphs is largely insensitive to this parameter (Fig.~\ref{fig:LFR_wl}), with ARI remaining stable for walks between 5 and 100 steps and peaking around 10–50. In contrast, Node2Vec performance dropped more at longer walks, likely due to over-smoothing. This stability makes TopKGraphs easier to tune in practice, especially in an unsupervised setting. Furthermore, we observed that TopKGraphs requires fewer walks to converge to the desired outcome, as illustrated in Fig.~\ref{fig:SBM_niter}.

\begin{figure}[h]  
    \centering
    \includegraphics[width=0.65\textwidth]{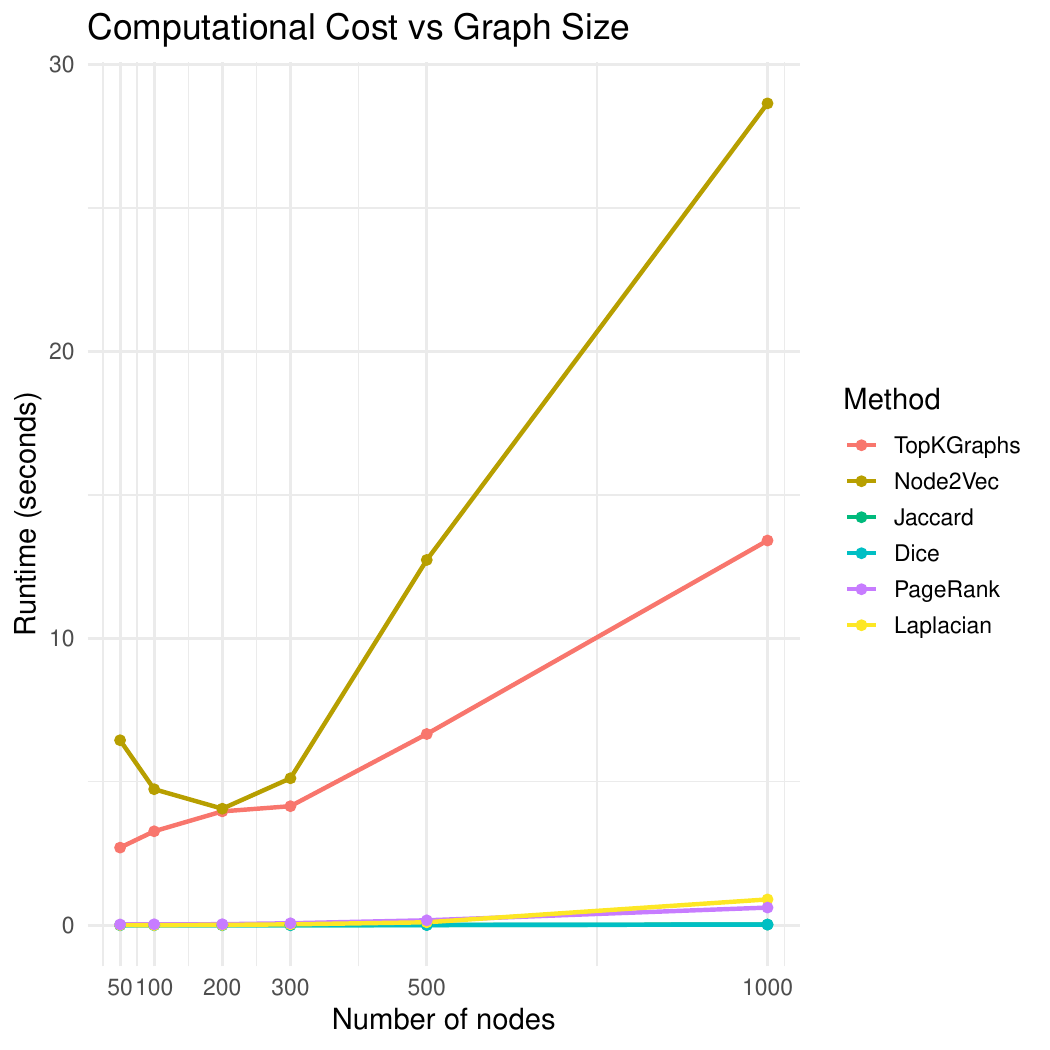}
    \caption{\textbf{Computational scaling of node-affinity methods on synthetic LFR graphs.}
Runtime (in seconds) as a function of graph size for six node-affinity approaches: TopKGraphs, Node2Vec, Jaccard similarity, Dice similarity, Laplacian embedding, and personalized PageRank. For each graph size (50–1000 nodes), an LFR benchmark graph was generated (average degree 5, maximum degree 10, mixing parameter $\mu = 0.05$, $\tau_1 = 2$, $\tau_2 = 1.1$), and the computation time required to obtain the corresponding affinity matrix or embedding was recorded. The figure illustrates the computational scaling behavior of the different methods as network size increases.}
    \label{fig:LFR_speed}
\end{figure}

Figure~\ref{fig:LFR_speed} summarizes runtime versus graph size on LFR instances. Node2Vec incurred the highest computational cost and scaled steeply with the total number of nodes. TopKGraphs was substantially faster than Node2Vec while being slower than single-pass similarities (Jaccard/Dice) and PageRank. Laplacian embedding remained efficient in this size regime. Overall, TopKGraphs offers a favorable accuracy--efficiency trade-off: near-Node2Vec accuracy at notably reduced runtime, and markedly higher accuracy than Jaccard/Dice/PageRank at moderate cost.

\subsubsection*{Tabular data as graphs: Breast Cancer Wisconsin}
\begin{figure}[h]  
    \centering
    \includegraphics[width=0.75\textwidth]{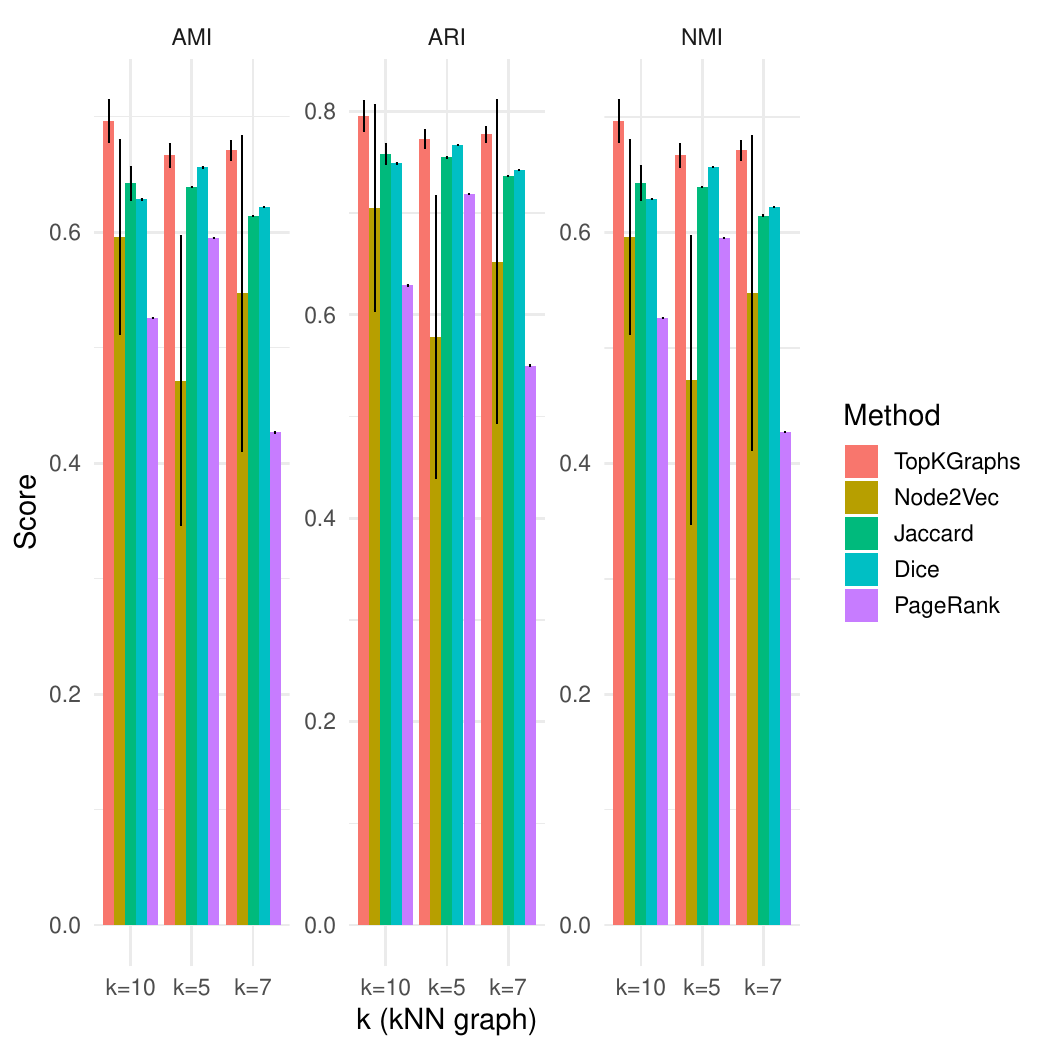}
    \caption{\textbf{Clustering performance on the Breast Cancer Wisconsin dataset using kNN graph representations.}
Bar plots show the mean clustering performance (± standard deviation over 50 runs) measured by Adjusted Rand Index (ARI), Normalized Mutual Information (NMI), and Adjusted Mutual Information (AMI). A k-nearest neighbor (kNN) graph was constructed from standardized feature vectors (30 features), with $k \in \{5,7,10\}$. Hierarchical clustering (Ward’s method) was applied to node-affinity matrices derived from TopKGraphs, Jaccard similarity, Dice similarity, personalized PageRank, and Node2Vec embeddings. Performance was evaluated against the ground-truth diagnostic labels (malignant vs. benign).}
    \label{fig:breast_cancer}
\end{figure}

On kNN graphs ($k \in {5,7,10}$) constructed from standardized tabular features, TopKGraphs achieved the best performance (Fig.~\ref{fig:breast_cancer}). Gains over Jaccard/Dice and PageRank were consistent across all metrics, highlighting the benefit of anchored random walks over pairwise overlap (Jaccard/Dice) and global diffusion (PageRank) when direct neighborhood structure is informative. This is also reflected by the fact that Jaccard and Dice showed better performance than PageRank and Node2Vec for this specific set-up.

\subsubsection*{Citation and protein interaction networks: CORA and PPI}
\begin{figure}[h]  
    \centering
    \includegraphics[width=0.7\textwidth]{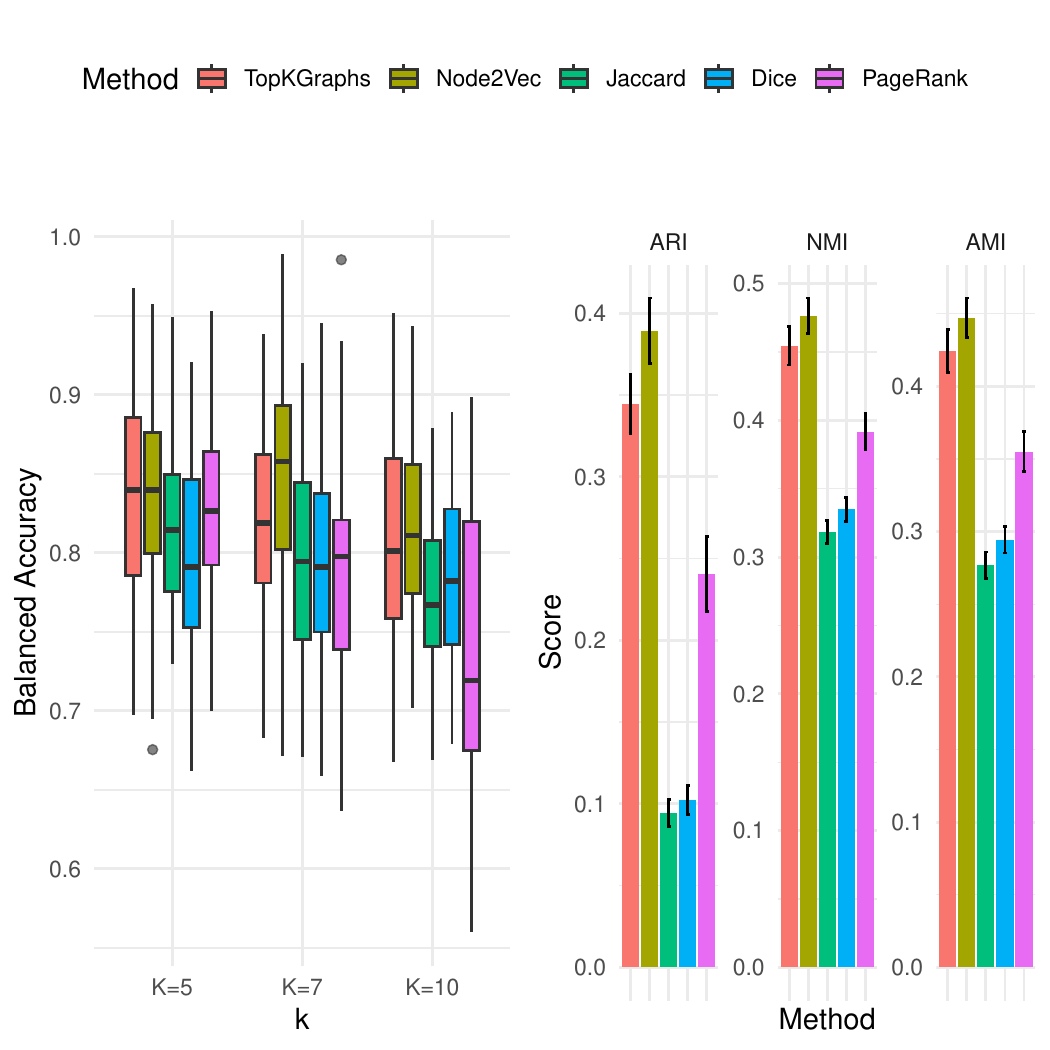}
    \caption{\textbf{Node-affinity evaluation on the CORA citation network.}
Performance was assessed on randomly sampled connected subgraphs (100 nodes) from the largest connected component of the CORA citation network over 50 runs. Node-affinity matrices were computed using TopKGraphs, Node2Vec, Jaccard similarity, Dice similarity, and personalized PageRank.
\emph{Left:} k-nearest neighbor (kNN) classification performance measured by Balanced Accuracy for $k \in \{5,7,10\}$, using distances derived from each affinity representation. Boxplots summarize variability across runs.
\emph{Right:} Community detection performance obtained via hierarchical clustering (Ward’s method) applied to each affinity matrix. Bars show mean ± standard error for Adjusted Rand Index (ARI), Normalized Mutual Information (NMI), and Adjusted Mutual Information (AMI), evaluated against the ground-truth paper subject classes.}
    \label{fig:CORA}
\end{figure}

On CORA subgraphs, TopKGraphs achieved strong community recovery and competitive kNN classification (Balanced Accuracy) across $k \in \{5,7,10\}$ (Fig.~\ref{fig:CORA}). It consistently outperformed Jaccard/Dice and PageRank and was competitive with Node2Vec across both clustering and classification endpoints.

\begin{figure}[h]  
    \centering
    \includegraphics[width=0.75\textwidth]{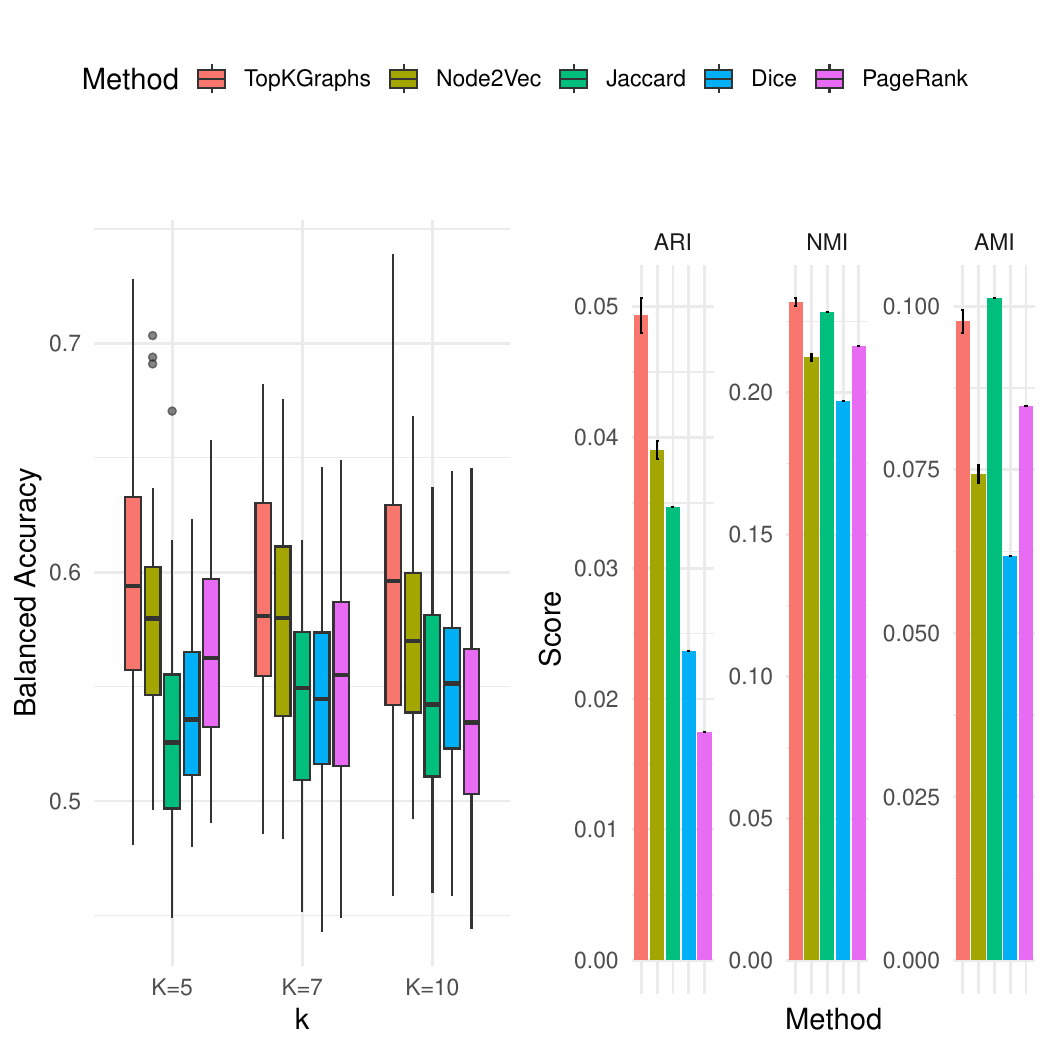}
    \caption{\textbf{Node-affinity evaluation on the human protein–protein interaction (PPI) network.}
Performance was assessed on randomly sampled connected subgraphs (100 nodes) from the largest connected component of a high-confidence STRING PPI network (combined score greater than 990), restricted to genes with curated disease annotations from DisGeNET. Small disease categories ($<20$ genes) and overly dominant classes (e.g., prostate cancer) were removed to obtain balanced disease communities. Results are aggregated over 50 runs. Node-affinity matrices were computed using TopKGraphs, Node2Vec, Jaccard similarity, Dice similarity, and personalized PageRank. \textit{Left:} k-nearest neighbor (kNN) classification performance measured by Balanced Accuracy for $k \in \{5,7,10\}$, using distances derived from each affinity representation. Boxplots summarize variability across runs.\textit{Right:} Community detection performance obtained via hierarchical clustering (Ward’s method) applied to each affinity matrix. Bars show mean ± standard error for Adjusted Rand Index (ARI), Normalized Mutual Information (NMI), and Adjusted Mutual Information (AMI), evaluated against disease-based gene annotations.}
    \label{fig:PPI}
\end{figure}

On the curated protein-protein interaction (PPI) network, the relative performance ordering differed between community detection and node classification tasks (Fig.~\ref{fig:PPI}). For hierarchical clustering, simple neighborhood overlap measures such as Jaccard similarity achieved competitive performance, in some cases approaching that of TopKGraphs. This indicates that disease-associated gene modules in the filtered PPI network exhibit substantial local neighborhood overlap, making purely local similarity sufficient for recovering coarse-grained community structure.

In contrast, a clearer separation between methods emerged in the k-nearest neighbor (kNN) classification setting. Here, TopKGraphs consistently achieved higher Balanced Accuracy across $k \in \{5,7,10\}$, while Jaccard and Dice similarities showed noticeably reduced performance. This divergence highlights an important distinction between clustering and local classification tasks: while clustering benefits from global consistency of pairwise similarities, kNN classification depends critically on the quality of local neighborhood rankings around individual nodes. In sparse and noisy PPI graphs, direct neighborhood overlap alone may be insufficient to reliably identify the most relevant disease-associated neighbors for a given gene.

Overall, TopKGraphs bridges the gap between local overlap and global diffusion. By biasing random-walk transitions toward proteins with similar neighborhoods to the start node, the method preserves the strong local signal captured by Jaccard similarity while enabling controlled multi-hop propagation. Aggregating first-visit rankings across multiple stochastic walks further stabilizes node affinities, yielding representations that are robust for both community detection and local classification. This consistency across evaluation tasks is particularly desirable in biological applications, where downstream objectives may range from module discovery to gene prioritization for individual disease contexts. Because affinities are derived from ranked visitation order rather than latent embeddings, TopKGraphs allows direct inspection of which proteins are prioritized for a given disease gene, facilitating biological interpretation and hypothesis generation.

\section{Conclusion}
\label{sec:conclusion} 
Across synthetic and real-world graphs, TopKGraphs consistently matches or outperforms strong baselines in both community detection and downstream classification, particularly in sparse, noisy, or weakly clustered networks. It provides a practical balance between accuracy, robustness, and computational efficiency, while its rank-based affinity matrices support direct inspection of node relationships, facilitating interpretability and hypothesis generation. Although TopKGraphs introduces additional computation relative to single-pass similarity measures, it remains substantially faster than Node2Vec in our experiments. Future work could explore optimized implementations and approximation strategies to scale the method to very large networks, as well as hybrid approaches that integrate TopKGraphs affinities with graph learning or embedding techniques to further enhance performance. In addition, convex optimization frameworks for latent signal reconstruction from multiple ranked lists \cite{schimek2024effective} could be explored to replace Borda aggregation, potentially yielding more stable and theoretically grounded consensus similarities under an additive noise model.

\backmatter





\bmhead{Acknowledgements}
The authors used ChatGPT (OpenAI) to improve the clarity and grammar of the manuscript. The authors reviewed and edited the content and take full responsibility for the final text.


\section*{Declarations}


\begin{itemize}
\item \textbf{Funding} \\
Not applicable.

\item \textbf{Conflict of interest/Competing interests}\\ 
There are no competing interests declared.

\item \textbf{Ethics approval and consent to participate} \\
Not applicable.
\item \textbf{Consent for publication} \\
Not applicable, as the data used in this study are fully anonymized and publicly available, and no individual person’s data are identifiable.
\item \textbf{Data availability} \\ 
All data, synthetic and data used for the applications are available on GitHub (\url{https://github.com/pievos101/TopKGraphs}). 
\item \textbf{Materials availability} \\
Not applicable.
\item \textbf{Code availability} \\
The proposed methods implemented within the R-package \texttt{TopKGraphs}, freely available on GitHub (\url{https://github.com/pievos101/TopKGraphs}).
\item \textbf{Author contribution} \\
BP developed the methods and conducted the simulation study. BP and MGS wrote the manuscript. All authors reviewed the manuscript.
\end{itemize}

\clearpage
\bibliography{sn-bibliography}

\end{document}